\documentclass{article}

\usepackage{nac_preprint}
\usepackage{float} 

\usepackage[T1]{fontenc}
\usepackage[utf8]{inputenc}
\usepackage{nicefrac}       
\usepackage{microtype}      

\usepackage{amsfonts}       
\usepackage{amsmath}
\usepackage{amssymb}
\usepackage{mathtools}
\usepackage[toc,page]{appendix}
\usepackage{enumitem}
\usepackage{graphics}
\usepackage{graphicx}
\usepackage{xcolor}
\usepackage{subcaption}

\usepackage[export]{adjustbox}
\usepackage{algorithm}
\usepackage{algorithmicx}
\usepackage[noend]{algpseudocode}
\algnewcommand{\LineComment}[1]{\State \(//\) #1}
\algnewcommand{\RLineComment}[1]{\State \(\triangleright\) #1}

\newcommand{\rulesep}{\unskip\ \hrule\ }
\usepackage{etoolbox}
\usepackage{tikz}
\usetikzlibrary{tikzmark}
\usetikzlibrary{calc}

\errorcontextlines\maxdimen

\newcommand{\ALGtikzmarkcolor}{black}
\newcommand{\ALGtikzmarkextraindent}{4pt}
\newcommand{\ALGtikzmarkverticaloffsetstart}{-.5ex}
\newcommand{\ALGtikzmarkverticaloffsetend}{-.5ex}
\makeatletter
\newcounter{ALG@tikzmark@tempcnta}

\newcommand\ALG@tikzmark@start{%
    \global\let\ALG@tikzmark@last\ALG@tikzmark@starttext%
    \expandafter\edef\csname ALG@tikzmark@\theALG@nested\endcsname{\theALG@tikzmark@tempcnta}%
    \tikzmark{ALG@tikzmark@start@\csname ALG@tikzmark@\theALG@nested\endcsname}%
    \addtocounter{ALG@tikzmark@tempcnta}{1}%
}

\def\ALG@tikzmark@starttext{start}
\newcommand\ALG@tikzmark@end{%
    \ifx\ALG@tikzmark@last\ALG@tikzmark@starttext
    \else
        \tikzmark{ALG@tikzmark@end@\csname ALG@tikzmark@\theALG@nested\endcsname}%
        \tikz[overlay,remember picture] \draw[\ALGtikzmarkcolor] let \p{S}=($(pic cs:ALG@tikzmark@start@\csname ALG@tikzmark@\theALG@nested\endcsname)+(\ALGtikzmarkextraindent,\ALGtikzmarkverticaloffsetstart)$), \p{E}=($(pic cs:ALG@tikzmark@end@\csname ALG@tikzmark@\theALG@nested\endcsname)+(\ALGtikzmarkextraindent,\ALGtikzmarkverticaloffsetend)$) in (\x{S},\y{S})--(\x{S},\y{E});%
    \fi
    \gdef\ALG@tikzmark@last{end}%
}

\apptocmd{\ALG@beginblock}{\ALG@tikzmark@start}{}{\errmessage{failed to patch}}
\pretocmd{\ALG@endblock}{\ALG@tikzmark@end}{}{\errmessage{failed to patch}}
\makeatother

\title{Convolutional Neural Generative Coding:\\ Scaling Predictive Coding to Natural Images}
 
\author{%
  Alexander Ororbia \\
  Rochester Institute of Technology \\
  \texttt{ago@cs.rit.edu}
  \And
  Ankur Mali\\
  University of South Florida\\
  \texttt{ankurarjunmali@usf.edu}
}

\begin{document}

\maketitle

\begin{abstract}
In this work, we develop \emph{convolutional neural generative coding} (Conv-NGC), a generalization of predictive coding to the case of convolution/deconvolution-based computation. Specifically, we concretely implement a flexible neurobiologically-motivated algorithm that progressively refines latent state feature maps in order to dynamically form a more accurate internal representation/reconstruction model of natural images. The performance of the resulting sensory processing system is evaluated on complex datasets such as Color-MNIST, CIFAR-10, and Street House View Numbers (SVHN). We study the effectiveness of our brain-inspired model on the tasks of reconstruction and image denoising and find that it is competitive with convolutional auto-encoding systems trained by backpropagation of errors and outperforms them with respect to out-of-distribution reconstruction (including the full 90k CINIC-10 test set).

\small{} 
\end{abstract}

\section{Introduction}
\label{sec:intro}

The algorithm known as backpropagation of errors \cite{bp2, linnainmaa1970representation} (or backprop) has served as a crucial element behind the tremendous progress that has been made in recent machine learning research, progress which has been accelerated by advances made in computational hardware as well as the increasing availability of vast quantities of data. Nevertheless, despite reaching or surpassing human-level performance on many different tasks ranging from those in computer vision \cite{he2015delving} to game-playing \cite{silver2016mastering}, the field still has a long way to go towards developing artificial general intelligence.  
In order to increase task-level performance, the size of deep networks has increased greatly over the years, up to hundreds of billions of synaptic parameters as seen in modern-day transformer networks \cite{floridi2020gpt}. However, this trend has started to raise concerns related to energy consumption \cite{patterson2021carbon} and as to whether such large systems can attain the flexible, generalization ability of the human brain \cite{brown2020language}. Furthermore, backprop itself imposes additional limitations beyond its long-argued biological implausibility \cite{crick1989recent, gardner1993neurobiology, shepherd1990significance}, such as its dependence on a global error feedback pathway for determining each neuron's individual contribution to a deep network's overall performance~\cite{minsky1961steps}, resulting in sequential backward, non-local updates that make parallelization difficult (which stands in strong contrast to how learning occurs in the brain \cite{jaderberg2016decoupled,ororbia2018conducting,ororbia2018biologically}).  
The limitations imposed by the prohibitively large size of these systems as well as the constraints imposed by their workhorse training algorithm, backprop, have motivated the investigation and development of alternative methodology. 

Some of the most promising pathways come from the emerging domain of research known as brain-inspired computation, which seeks to develop neural architectures and their respective credit assignment algorithms that leverage only local information, motivated strongly by how learning is conducted by the brain. The promise of brain-inspired computing brings with it synaptic adjustment mechanisms that are neurobiologically-grounded \cite{hebb1949organization} as well as neural computation and inference that is flexible, capable of conducting a wide variety of operations \cite{ororbia2022neural} at biologically more faithful levels \cite{maass1997networks,ororbia2019spiking}, facilitating massive algorithmic parallelization (at scale) and adaption on analog and neuromorphic hardware \cite{furber2016large,roy2019towards,kendall2020training}. 

In addressing the many-fold challenges facing backprop-based ANNs and in the direction of brain-inspired computing, we design a new model for image processing, convolutional neural generative coding (Conv-NGC), which is inspired by human learning. It is well-known that humans process information with hierarchical top-down feed-forward and bottom-up backward connections, which continuously \emph{predict and correct} humans' internal representations of that information~\cite{rao1999predictive}. In neuroscience, this interplay is known as predictive coding~\cite{rao1999predictive,rainer1999prospective}. 
Similarly, Conv-NGC includes state prediction and correction steps that continuously generate and refine its internal representations (Figure \ref{fig:architecture}). Furthermore, Conv-NGC encodes complex visual information by incorporating blocks of (de)convolutional into a top-down directed generative model within the framework of predictive coding \cite{bastos2012canonical,chalasani2015contextcdn,clark2015surfing,ororbia&mali2019lifelong,ororbia2022neural}. 

Our contributions are as follows: 
\textbf{1)} We propose a new neural perception model, Conv-NGC, which acquires robust representations of natural images in an unsupervised fashion, 
\textbf{2)} To the best of our knowledge, this is the first work in the literature where visual inputs are processed using deep (de)convolutional layers directly and naturally within of  framework of predictive coding, significantly enhancing its representation power for vision-based tasks, and, 
\textbf{3)} We demonstrate, on several natural image datasets, that the proposed Conv-NGC is competitive with existing backprop-based models (of similar architectural designs) on the tasks of image reconstruction and image denoising and outperforms them with respect to out-of-distribution prediction.

\section{Convolutional Neural Generative Coding}
\label{sec:model_details}

We start by describing our model instantiation of convolutional neural generative coding (Conv-NGC), which is tasked with learning from streams of natural images in an unsupervised fashion. Typically, in a visual recognition task, input patterns belonging to different object classes, arranged into batches, are presented to a processing system at different time points for training. In this section, we describe our problem setup and model architecture (see the Appendix for related work and a detailed discussion of neurobiological motivations). 


\begin{figure}
  \begin{center}
    \includegraphics[width=0.75\textwidth]{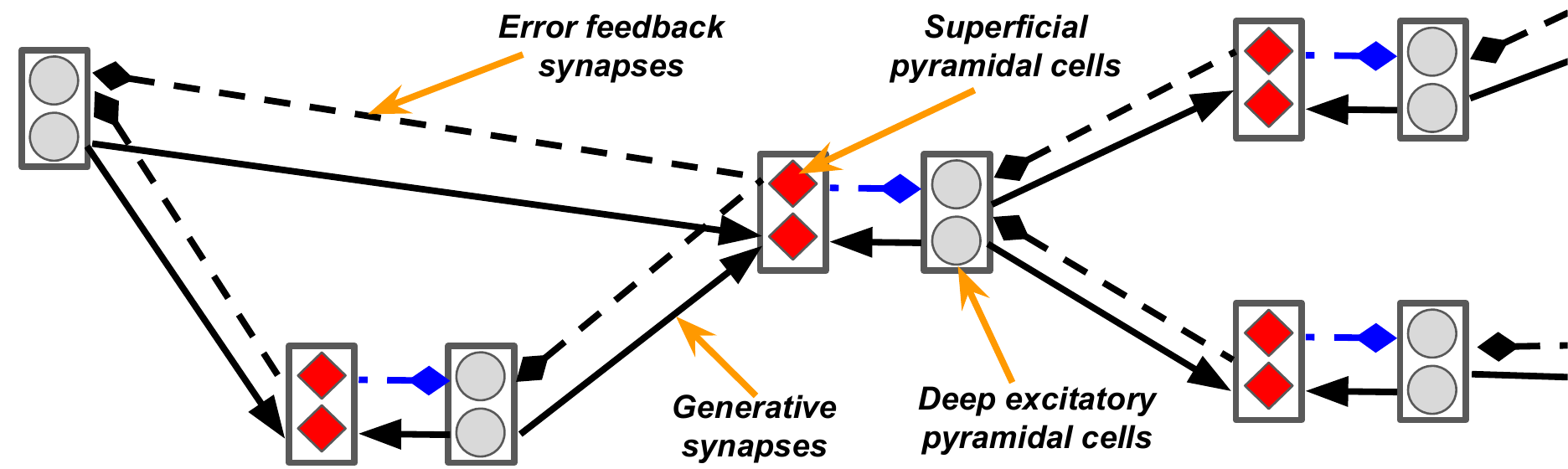}
  \end{center}
  \vspace{-0.15cm}
  \caption{An arbitrary NGC circuit. Grey circles are state units (deep excitatory pyramidal cells) and red diamonds are error units (superficial pyramidal cells). Solid arrows are predictive (generative) synapses and dashed diamond ones are error (feedback) synapses.}
  \label{fig:ngc_circuit}
  \vspace{-0.5cm}
\end{figure}

\noindent 
\textbf{Problem Setup: } With respect to the problem setup, a Conv-NGC system is tasked with processing a finite collection of images depicting a certain set of object classes arranged in an arbitrary order. The dataset contains a set of $n$ input samples: $\mathcal{D}=\{(\mathbf{x}_1,\mathbf{y}_1 \dots (\mathbf{x}_{n},\mathbf{y}_{n})\}$. Here, $\mathbf{x}_j \in \mathcal{R}^{I \times I \times C}$ represents the image of the $j^\text{th}$ input sample ($I \times I$ is the 2D shape of any single image channel and $C$ is the number of channels, i.e., $C=3$ for the red-green-blue channels) and $\mathbf{y}_j \in \{0,1\}^{Y \times 1}$ is its ground truth class label ($Y$ is the number of classes). Note that while we formalize the labels available in each benchmark dataset, the models that we study in this work are unsupervised and, as a result, never make use of the labels in $\mathcal{D}$.

\noindent
\textbf{Notation:} In this study, the symbol $*_s$ is used to refer to a strided convolution where $s$ is the stride argument ($*_1$ means convolution with stride of 1, which would also be the same as just $*$). In contrast, the symbol $\circlearrowleft_s$ denotes deconvolution (or transposed convolution) with a stride of $s$. The Hadamard product is denoted by $\odot$ while $\cdot$ represents a matrix/vector multiplication. $()^T$ denotes the transpose operation. 
$\text{Flatten}(\mathbf{z})$ means that the input tensor $\mathbf{z}$ is converted to a column vector with a number of rows equal to the number of elements that it originally contained while $\text{UnFlatten}(\mathbf{z})$ is its inverse (i.e., it converts the vector back to its original tensor shape). Finally, $\text{Dilate}(\mathbf{v}, s)$ is used to represent a dilation function controlled by the dilation (integer) size (e.g., $s = 2$). Note that Conv-NGC is technically made up of 4D synaptic tensors, thus, when we write $\mathbf{W}_{ij}$, we are saying that we are retrieving a 2D matrix at position $(i,j)$ in the 4D tensor $\mathbf{W}$ (for extracting a scalar in $\mathbf{W}$, one would write $W_{ijkl}$, without bold font).


\subsection{Deep Convolutional Neural Coding}
\label{sec:algo}

Conv-NGC is a generalization of the NGC computational framework in \cite{ororbia2022neural} to the case of natural image data. The underlying process of Conv-NGC can be divided into three components: 
\textbf{1)} local prediction and error unit map calculation, 
\textbf{2)} latent state map correction, and 
\textbf{3)} local synaptic adjustment. 
See Figure \ref{fig:ngc_circuit} for a depiction of a general NGC circuit (and see Figure \ref{fig:architecture} for Conv-NGC).

\noindent 
\textbf{The Neural Coding Process: }
Fundamentally, Conv-NGC consists of a set of $L$ predictive layers (typically arranged hierarchically, though this is not a strict architectural requirement, e.g., Figure \ref{fig:ngc_circuit} depicts a non-hierarchical NGC circuit) that are designed to learn latent representations of observed patterns. It is important to note that, unlike the bottom-up forward propagation of a standard convolutional neural network (CNN), neural layers within a Conv-NGC system make top-down predictions, the errors of which are then used to subsequently correct the layers' own values. In effect, this means that the layers in Conv-NGC are stateful and their computation within a forward pass can be further broken down into distinct computations -- top-down prediction and state correction. 

First, in the top-down prediction phase, given its current state $\mathbf{z}^{\ell}$ (which abstractly models the functionality of deep excitatory pyramidal cells), each layer $\ell$ of our model tries to predict the state of the layer below it, yielding prediction $\mathbf{\bar{z}}^{\ell-1}$. At the bottom-most layer, the model predicts $\mathbf{\bar{z}}^{0}_{x}$ for the input data pattern ($\mathbf{x}$). 
Following this, a set of error neurons (which abstractly model the functionality of superficial pyramidal cells) compute the mismatch between this prediction and the actual state $\mathbf{z}^{\ell-1}$, i.e., $\mathbf{e}^{\ell-1} = (\mathbf{z}^{\ell-1} - \mathbf{\bar{z}}^{\ell-1})$.

Second, in the correction phase, the model's internal state layers are corrected based on how accurate their top-down guesses turned out to be. The error/mismatch signal $\mathbf{e}^{\ell-1}$ computed during the prediction phase is subsequently used to adjust the current values of the state $\mathbf{z}^\ell$ that originally made the prediction $\mathbf{\bar{z}}^{\ell-1}$. This \emph{local error correction}, which is not present in traditional feedforward ANNs, helps to nudge the state $\mathbf{z}^\ell$ towards a configuration (i.e., set of values) that better predicts the layer below in the future and thus moves the layer towards a better representation/higher-level abstraction of the input. This correction is ``local'' in  the sense that each layer's update depends only on a top-down error signal, which is produced by comparing its own values with the predictions made by the layer above it, and a bottom-up error signal, which is produced by comparing its predictions of a nearby layer's activity values and that state's current actual values.

Given the description of the two general computations above, we may now describe how Conv-NGC processes data. Unlike a standard feedforward ANN, which predicts 
$\mathbf{y}_j$ given $\mathbf{x}_j$ (or, in the case of auto-encoding, the ANN attempts to reconstruct $\mathbf{x}_j$ itself given $\mathbf{x}_j$ as input) with a single forward pass, our model works in multiple steps. 
First, a Conv-NGC network predicts the value of $\mathbf{z}^0 = \mathbf{x}_j$ from $\mathbf{z}^1$ (which generates prediction $\mathbf{\bar{z}}^0$), $\mathbf{z}^1$ from $\mathbf{z}^2$ (which generates prediction $\mathbf{\bar{z}}^1$), etc. As each prediction is made, the corresponding set of error neurons compute the mismatch between the predicted value and its target state layer.\footnote{These predictions and error neuron calculations, although presented as occurring sequentially, can naturally be made in parallel given that they depend on immediately/locally available values.} 
Next, the neural system then corrects the values of its states $\{\mathbf{z}^1, \mathbf{z}^2, \cdots, \mathbf{z}^L\}$ given the current values of the error neurons. 
These two steps are then repeated for several iterations, specifically over a stimulus window of $T$ steps, to arrive at a set of internal representations that accurately represent the input. This means that each layer/region $\ell$ of a neural coding model tries to satisfy two main objectives: 
1) to uncover a better latent representation in order to predict a nearby neural region/layer (a bottom-up adjustment), and 
2) to be closer to what the layer above ($\mathbf{z}^{l+1}$) predicted its state should be (a top-down expectation). 
By performing several iterations of top-down prediction followed by state correction, the model not only minimizes layer-specific predictions but also optimizes its global (system) representation for the current dataset. Implementation details of the above processes, specifically generalized to the case of feature maps and (de)convolution, are provided in the next section.

\section{Neural Coding Training and Inference}
\label{sec:training_eval}

\begin{figure*}[!t]
\begin{center}
\includegraphics[trim=0 0 5 0, clip,width=0.75\textwidth]{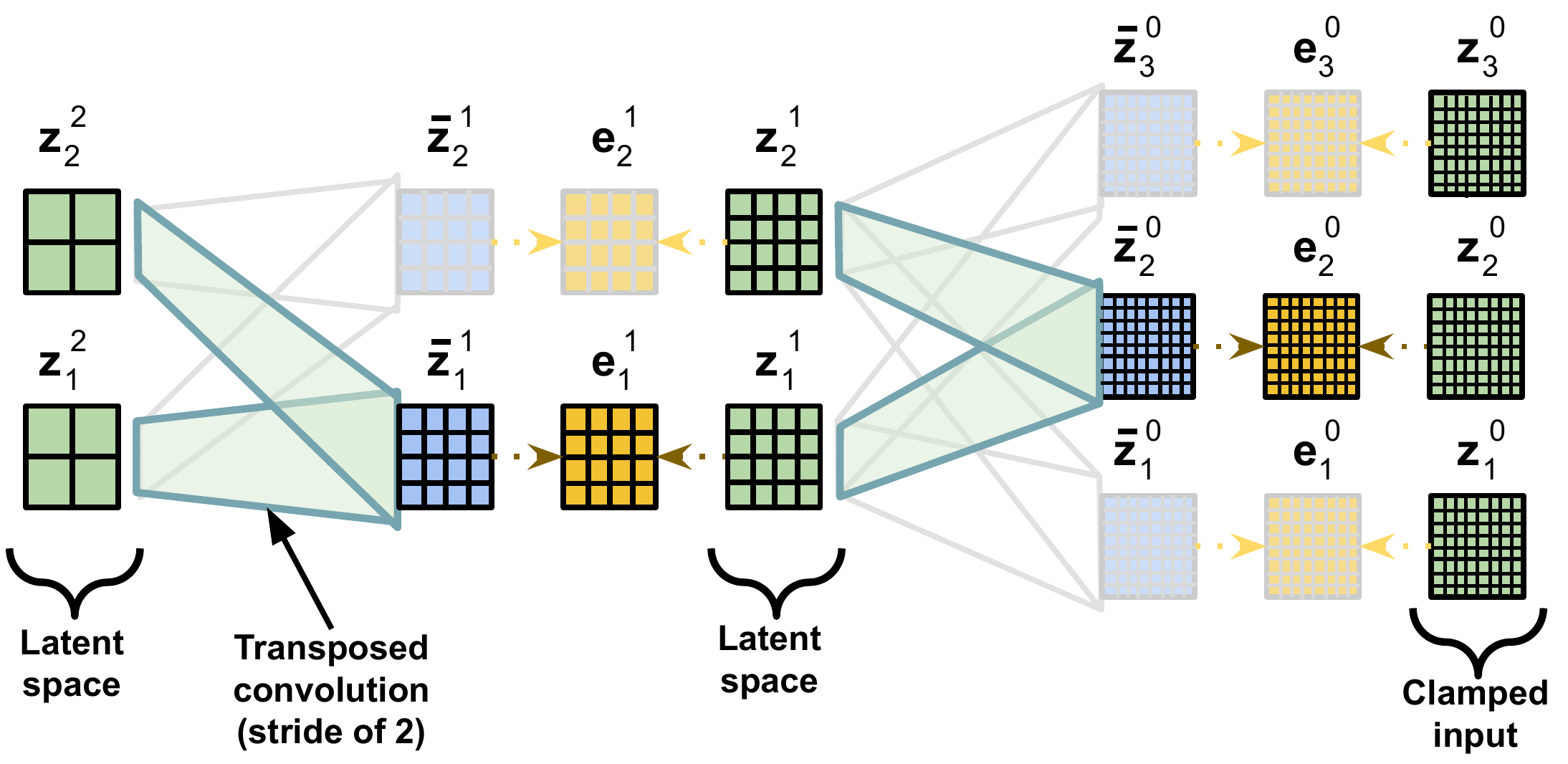}
\caption{A 3-layer convolutional neural generative coding network. Each latent state map $\mathbf{z}^\ell_i$ generates a prediction of each latent map $\mathbf{z}^{\ell-1}_j$ in the layer below via deconvolution. In non-transparent colors, a possible prediction pathway within the circuit is shown (transparent colors indicate no involvement in this pathway), i.e., state maps $\mathbf{z}^2_1$ and $\mathbf{z}^2_2$ each contribute to the prediction $\mathbf{\bar{z}}^1_1$ of map $\mathbf{z}^1_1$ while state maps $\mathbf{z}^1_1$ and $\mathbf{z}^1_2$ each contribute to the prediction $\mathbf{\bar{z}}^0_2$ of map $\mathbf{z}^0_2$ (note: in the bottom layer, each $\mathbf{z}^0_j$ map could represent a particular color channel of an input image). Error unit map $\mathbf{e}^1_1$ encodes the mismatch between prediction $\mathbf{\bar{z}}^1_1$ and the value of latent map $\mathbf{z}^1_1$ while error unit map $\mathbf{e}^0_2$ encodes the mismatch between prediction $\mathbf{\bar{z}}^0_2$ and the value of latent map $\mathbf{z}^0_2$. Green squares depict latent unit maps, blue squares depict prediction output maps, and orange/yellow squares depict error unit maps.}
\label{fig:architecture}
\end{center}
\end{figure*}

In this section, we provide concrete implementation details of the neural coding process described earlier, depicting how layer-wise state prediction, state-correction, and synaptic parameter updating occur specifically in the context of visual object reconstruction. 
The first two steps iteratively predict and correct the representations of the Conv-NGC model for observed input values (natural images) of the current dataset. After $T$ iterations, the final step entails adjusting model synaptic efficacies using simple Hebbian-like updates. 
We start by providing the mechanics of the three above steps and then describe the objective function that our model dynamically optimizes. The full algorithmic specification of Conv-NGC is presented in the pseudocode in the Appendix 
and an example 3-layer Conv-NGC model is visually depicted in Figure \ref{fig:architecture}. 

\subsection{Inference: Predicting and Correcting Neural States} 
\label{subsec:prediction}
For each layer of a Conv-NGC system, note that the full state of any given layer $\ell$ is represented by a set of $C_\ell$ feature state maps, i.e., $\{ \mathbf{z}^\ell_1, \mathbf{z}^\ell_2,...,\mathbf{z}^\ell_{C_\ell} \}$ where each state map $\mathbf{z}^\ell_i$ is essentially a block/cluster of neurons (that encode a partial distributed representation of the detected input below), meaning that any layer consists of $C_\ell$ channels. To initialize each state map, instead of setting each to be a grid of zero values (similar to how neural vectors are initialized in fully-connected predictive coding models \cite{ororbia2022neural}), we initialize the states with a top-down ancestral projection pass by first sampling a random noise value for the top-most set of latent state maps, i.e., $\mathbf{z}^L_c \sim \mathcal{N}(\mu_z,\sigma_z), \; c = 1,...,C_L$ (we set $\mu_z = 5$ and $\sigma_z = 0.05$ in this work), and then project this sampled state down along the Conv-NGC network to obtain the initial values of the other layers (see Appendix for details). 

\noindent
\textbf{Layer-wise State Map Prediction: } 
At each layer, the $i$th feature map state $\mathbf{z}^{\ell}_i$\footnote{A complete vector representation of a neural layer's entire state is the concatenation of all of its $C$ (flattened) feature state maps, i.e., $\mathbf{z}^\ell = [\mathbf{z}^{\ell}_1, \mathbf{z}^{\ell}_2, \cdots,\mathbf{z}^{\ell}_C] $.} is used to (partially) predict the $j$th feature map state of the layer below it, producing the prediction $\mathbf{\bar{z}}^{\ell-1}_j$. These local layer-wise predictors perform their computation independently (in parallel) and are coordinated through error units at each layer $\ell$ for each feature map. Specifically, the error neurons $\mathbf{e}^{\ell-1}_j$ at layer $\ell-1$ (for the $j$th channel) compute the difference between the prediction $\mathbf{\bar{z}}^{\ell-1}_j$ (from the layer $\ell$ above) and the target map activity $\mathbf{z}^{\ell-1}_j$. This error message, in turn, is used to (partially) adjust the state representation $\mathbf{z}^\ell$ at layer $\ell$. Formally, the predictor and error neurons 
are computed on a per-feature map basis as follows: 
\begin{align}
    \mathbf{\bar{z}}^{\ell-1}_j = g^\ell\Big( \Big( \sum^{C_{\ell}}_i \mathbf{W}^\ell_{ij} \circlearrowleft_s \phi^\ell( \mathbf{z}^{\ell}_i ) \Big) + \mathbf{b}^\ell_j \Big),\;  \mathbf{e}^{\ell-1}_j = (\mathbf{z}^{\ell-1}_j - \mathbf{\bar{z}}^{\ell-1}_j), \label{eqn:prediction}
\end{align}
where $\phi^\ell()$ is the nonlinear activation/transfer activation applied to the state activity map $\mathbf{z}^\ell_i$, $\mathbf{W}^\ell_{ij}$ is the $j$th learnable kernel for the $i$th input state map, and $\mathbf{b}_j$ is the $j$th bias map applied to the prediction/output map $\mathbf{\bar{z}}^\ell_j$. $g^{\ell}()$ is the predictive output nonlinearity applied to the result $\mathbf{\bar{z}}^\ell_j$ (typically chosen according to the distribution that is chosen to model the state of $\ell-1$, e.g., the identity for a multivariate Gaussian model). 
Note that in the above equation, in a (de)convolutional layer, a final complete prediction of the $j$th output channel involves the summation/aggregation of multiple filters applied to each (input) state feature map of layer $\ell$ (up to $C_\ell$ input channels). Extending the prediction operation of Figure \ref{eqn:prediction} to make use of stacks of operations, such as within a residual block, is detailed in the Appendix (which contains a biologically-plausible scheme for handling credit assignment for operation blocks).

\noindent
\textbf{State Map Correction: } 
During this step, the ($i$-th) state feature map $\mathbf{z}^\ell_i$ is refined using the values of the error neurons of the current and previous layers, i.e., $\ell$ and $\ell-1$. For the top-most layer, the error $\mathbf{e}^L_i$ of any feature map $i$ does not exist so it is not computationally modeled/simulated. 
Formally, the $\ell$-th predictor (or rather, its ($i$-th) map $\mathbf{z}^\ell_i$ ) corrects its state values using both the bottom-up and top-down error messages/signals according to the following (Euler) integration scheme below: 
\begin{equation}
    \begin{aligned}
    \mathbf{z}^\ell_i \leftarrow \big( \mathbf{z}^\ell_i + \beta \mathbf{d}^\ell_i - \gamma \mathbf{z}^\ell_i \big),  \text{where } \mathbf{d}^\ell_i = -\mathbf{e}^\ell_i +  \sum^{C_{\ell-1}}_j \mathbf{E}^\ell_{ji} *_s \mathbf{e}^{\ell-1}_j,
\end{aligned} \label{eq:correction}
\end{equation}
where $\leftarrow$ denotes a variable override and the modulation/adjustment factor $\beta$ is a constant value, which controls the state correction rate, and $-\gamma \mathbf{z}^\ell_i$ is the leak variable, controlled by the strength factor $\gamma$ (set to a small value such as $0.001$, also serving as a form of light regularization applied to the latent state map neurons). 
The (4D) error kernel tensor $\mathbf{E}^\ell$ models a learnable feedback pathway that is responsible for transmitting the error from the layer $\ell-1$ to the layer $\ell$. 
This means that the bottom-up error message is produced by aggregating across the $C_{\ell-1}$ output channels/state maps that make up layer $\ell-1$, i.e., a convolution must be applied to each output channel $j$ as follows: $\mathbf{E}^\ell_{ji} *_s \mathbf{e}^{\ell-1}_j$, meaning that the prediction of each lower-level state map that was (partially) made by the state map/predictor $\mathbf{z}^\ell_i$ contributes equally to its final value, resulting in a state map correction (which is further modulated by the top-down pressure $\mathbf{e}^\ell_i$,  exerted by the layer above $\mathbf{z}^{\ell+1}$). Although having separate error kernels for transmitting error message signals resolves the weight transport problem that characterizes backprop-based learning (and is thus more biologically plausible), this comes at an increased memory footprint and additional computation which is needed for updating the error filters themselves. Thus, we set the error filters, in this work, to be $\mathbf{E}^\ell_{ji} = (\mathbf{W}^\ell_{ij})^T$, to speed up simulation (with no  noticeable change in performance). 
Note that, for test time inference, we do not adjust synaptic efficacies and 
either clamp or initialize the bottom state $\mathbf{z}^0 = \mathbf{x}$. To obtain a prediction, the neural system will conduct $T$ steps of state prediction and correction, eventually outputting $\mathbf{\bar{z}}^0$.


\subsection{Training: Updating Model Parameters}
\label{subsec:update}

\noindent
\textbf{Neural Coding Synaptic Update: } After $T$ iterations of state prediction and correction (as in the previous section), the updates to each state prediction filter $\mathbf{W}^\ell_{ij}$ and error filter $\mathbf{E}^\ell_{ji}$ are computed according to a Hebbian-like update: 
\begin{align*}\label{eq:update_coding}
    \Delta {\mathbf{W}^\ell}_{ij} = \mathbf{e}^{\ell-1}_j *_{1} \text{Dilate}\Big( \big( \phi^\ell(\mathbf{z}^{\ell}_i) \big)^T, s \Big), \notag \\ \quad \Delta {\mathbf{E}^\ell}_{ji} =  \lambda \Big(  \text{Dilate}\Big( \big( \phi^\ell(\mathbf{z}^{\ell}_i) \big)^T, s \Big) *_1 \mathbf{e}^{\ell-1}_j \Big)
\end{align*}
where 
$\lambda$ is modulation factor meant to control the time scale of the evolution of the error filters (and generally set to $<1.0$, e.g., $0.9$) - note that this part of the update rule is discarded if $\mathbf{E}^\ell_{ji} = (\mathbf{W}^\ell_{ij})^T$ as we do in this work. 
The above local update rule is a generalization of the one proposed in \cite{ororbia&mali2019lifelong,ororbia2022neural} to the case of a (de)convolutional filter. After the update for a particular filter has been calculated and it has been used to adjust the current physical state of the kernel synapses, we further normalize/constrain each kernel such that its Euclidean norm does not exceed one (see the Appendix for the specification of the re-projection step). 
This constraint ensures that Conv-NGC avoids the degenerate solution of simply increasing its synaptic kernel values while obtaining only small/near-zero latent activity values, much as is done in convolutional sparse coding \cite{heide2015fast} (this also means that one could also view Conv-NGC as a sort of ``deep'' convolutional sparse coding).

\noindent 
\textbf{Objective Function: }
During training, a Conv-NGC model refines its internal states such that the output of the local predictions move as close as possible to the actual values of the state maps, which means that the bottom-most layer stays as close as possible to the sensory input $\mathbf{x}_j$. In order to do this, Conv-NGC optimizes \emph{total discrepancy optimization} (ToD)  \cite{ororbia2017learning,ororbia2018biologically,ororbia2022neural} via the prediction, state correction, and synaptic adjustment steps presented in the earlier sub-sections. The simplest version of this optimization function is defined as the sum of mismatches between predictions and actual states at each level of the model:
\begin{equation}\label{eqn:total_discrep}
    \mathcal{L}_{ToD} =  \sum_\ell \Big( -\frac{1}{2} || \mathbf{z}^\ell - \mathbf{\bar{z}}^\ell ||^p_q \Big), \mbox{ where, } p = q =  2. 
\end{equation}
The online minimization of the discrepancy among states as depicted in Figure \ref{eqn:total_discrep} progressively refines the representation of all layers in our model. Notably, the total discrepancy objective can also be viewed as approximately minimizing a form of (variational) free energy \cite{friston2010free}, representing a concrete statistical learning connection to and implementation of a prominent neuro-mechanistic Bayesian brain theory.

\section{Experiments}
\label{sec:experiments}

Given our specification of the Conv-NGC model's inference and learning processes, we next describe the experimental setup. 
We compare Conv-NGC to powerful backprop-based models such as the convolutional autoencoder (Conv-AE) and the denoising convolutional autoencoder (Conv-DAE) as well as the fully-connected form of neural generative coding (NGC), which under certain conditions, recovers the unsupervised form of the model in \cite{salvatori2021associative} (the hierarchical predictive coding network, or PCNN).  
With respect to the datasets used in our simulations, we utilized Color-MNIST, CIFAR-10, and SVHN (the Street View House Numbers database) to both train and evaluate the baselines and Conv-NGC (using respective training/validation/testing splits) while CINIC-10 was only used as an additional test set for the out-of-distribution reconstruction experiments. All datasets consisted of $32\times32$ complex natural images. In the Appendix, we provide baseline descriptions, configuration details of the Conv-NGC model simulated in this study, and dataset details.

\begin{table*}[htb!]
\caption{Model reconstruction and denoising performance on the test samples of Color-MNIST (top), CIFAR-10 (middle), and SVHN (bottom). Reported measurements are mean and standard deviation across five trials. (Note: model output values are re-scaled to between $[0, 255]$ when calculating metrics.) In terms of performance, a lower MSE and higher SSIM are better.}
\begin{center}
\begin{tabular}{c| c c | c c} 
\hline
 \textbf{Color-MNIST} & \multicolumn{2}{|c}{Reconstruction} & \multicolumn{2}{|c}{Denoising ($\sim \mathcal{N}(0,0.1)$)}\\
 \textbf{Model} & \textbf{MSE} & \textbf{SSIM} & \textbf{MSE} & \textbf{SSIM} \\ 
 \hline\hline
 Conv-AE & $32.99 \pm 1.0680$ & $0.9021 \pm 0.0100$ & $176.9321 \pm 2.0923$ & $0.6912 \pm 0.007$\\
 Conv-DAE & $23.09 \pm 0.4722$ & $0.9324 \pm 0.0060$ & $\mathbf{76.8610 \pm 0.8511}$ & $\mathbf{0.8424 \pm 0.003}$\\
 NGC & $328.5659 \pm 3.5000 $ & $0.7652 \pm 0.0032$ & $887.205 \pm 18.2500 $ & $0.5476 \pm 0.0075$ \\
 Conv-NGC & $\mathbf{11.2802 \pm 0.5008}$ & $\mathbf{0.9838 \pm 0.0005}$ & $202.2222 \pm 1.0511$ & $0.6732 \pm 0.0006$\\ 
\hline
 \textbf{CIFAR-10} & \multicolumn{2}{|c}{Reconstruction} & \multicolumn{2}{|c}{Denoising ($\sim \mathcal{N}(0,0.1)$)}\\
 \textbf{Model} & \textbf{MSE} & \textbf{SSIM} & \textbf{MSE} & \textbf{SSIM} \\ 
 \hline\hline
 Conv-AE & $13.6890 \pm 1.6320 $ & $0.9310 \pm 0.0030$ & $208.4608 \pm 9.3928$ & $0.7628 \pm 0.0003$\\
 Conv-DAE & $8.8810 \pm 1.5311 $ & $0.9720 \pm 0.0010$ & $\mathbf{16.7781 \pm 5.1037}$ & $\mathbf{0.9110 \pm 0.0020}$\\
 NGC & $413.5140 \pm 7.1000$ & $0.7230 \pm 0.0003$ & $913.2587 \pm 21.2000 $ & $0.5308 \pm 0.0025$\\
 Conv-NGC & $\mathbf{6.3668 \pm 1.2522}$  & $\mathbf{0.9955 \pm 0.0009}$ & $183.8624 \pm 9.4002$ & $0.8603 \pm 0.0007$ \\ 
\hline
 \textbf{SVHN} & \multicolumn{2}{|c}{Reconstruction} & \multicolumn{2}{|c}{Denoising ($\sim \mathcal{N}(0,0.1)$)}\\
 \textbf{Model} & \textbf{MSE} & \textbf{SSIM} & \textbf{MSE} & \textbf{SSIM} \\ 
 \hline\hline
 Conv-AE & $7.5043 \pm 2.9351 $  & $0.8934 \pm 0.0002$ & $154.6783 \pm 25.7555 $ & $0.7490 \pm 0.0030$\\
 Conv-DAE & $2.7263 \pm 1.9286 $ & $0.9510 \pm 0.0002$ & $\mathbf{87.9926 \pm 7.2091}$ & $\mathbf{0.9002 \pm 0.0030}$\\
 NGC & $67.393 \pm 9.02 $ & $0.9436 \pm 0.0028$ & $1116.764 \pm 283.55 $ & $0.6704 \pm 0.0904$ \\
 Conv-NGC & $\mathbf{1.9500 \pm 0.5609}$ & $\mathbf{0.9976 \pm 0.0002}$ & $97.2695 \pm 3.5245$ & $0.8650 \pm 0.0007$\\ 
 \hline
\end{tabular}
\end{center}

\label{table:recon_denoise}
\end{table*}


\noindent
\textbf{Reconstruction and Denoising:} In this set of tasks, we investigate each model's ability to reconstruct the given input images from each dataset as well as to recover the original image values under corruption noise. 
For the task of reconstruction, we measure the mean squared error (MSE) for each input between the model's predicted values $\mathbf{\hat{x}}_i$ and the original image $\mathbf{x}_i$. To better probe the image quality of reconstructed outputs, we evaluate the structural similarity index measure (SSIM). Finally, for the more complex natural images in SVHN and CIFAR-10, we measure peak signal-to-noise ratio (PSNR). Note that, in the Appendix, we present the mathematical formulas and details for each of these metrics. For the results of the reconstruction and denoising tasks, see Tables \ref{table:recon_denoise} and \ref{table:psnr_analysis}. 

\begin{figure}
     \centering
     \begin{subfigure}[b]{0.215\textwidth}
         \centering
         \includegraphics[width=\textwidth]{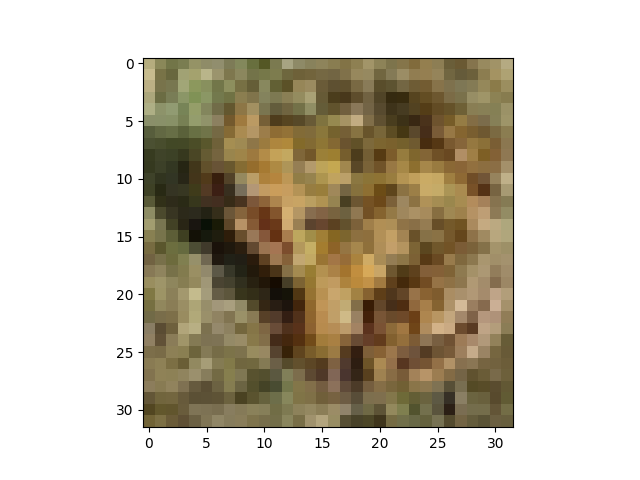}
         \caption{Original image.}
         \label{fig:cifar_orig}
     \end{subfigure}
     \begin{subfigure}[b]{0.215\textwidth}
         \centering
         \includegraphics[width=\textwidth]{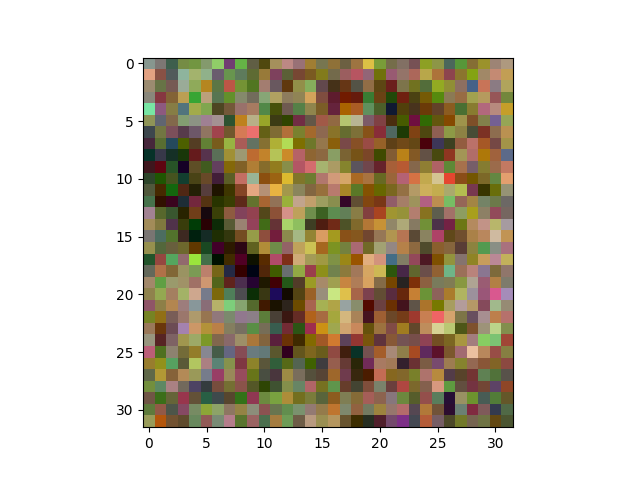}
         \caption{Corrupted image.}
         \label{fig:cifar_corrupt}
     \end{subfigure}
     \begin{subfigure}[b]{0.215\textwidth}
         \centering
         \includegraphics[width=\textwidth]{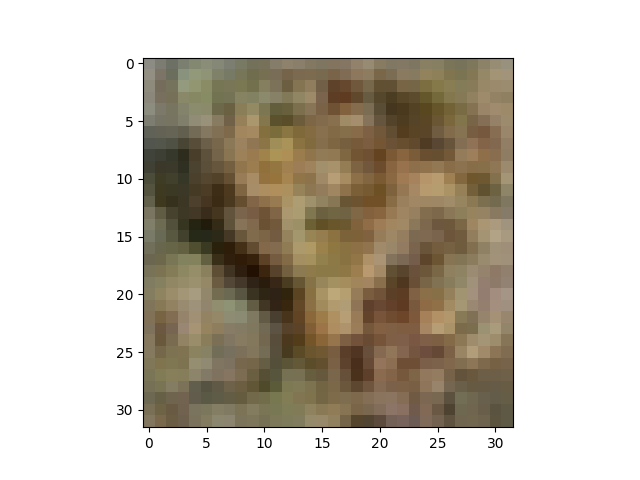}
         \caption{Denoised image.}
         \label{fig:cifar_recon}
     \end{subfigure}
    \\
     \begin{subfigure}[b]{0.215\textwidth}
         \centering
         \includegraphics[width=\textwidth]{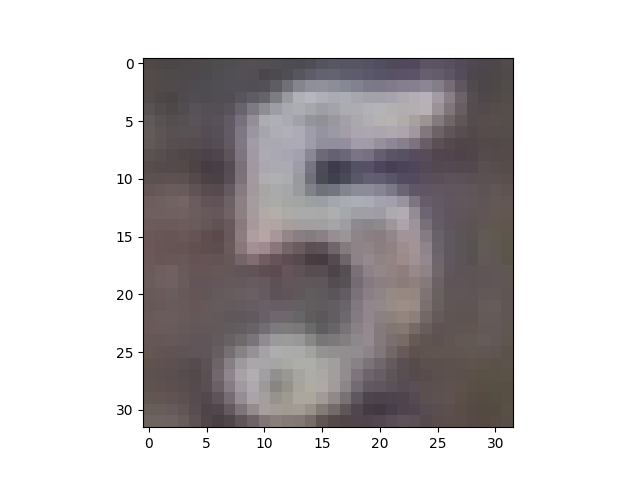}
         \caption{Original image.}
         \label{fig:svhn_orig}
     \end{subfigure}
     \begin{subfigure}[b]{0.215\textwidth}
         \centering
         \includegraphics[width=\textwidth]{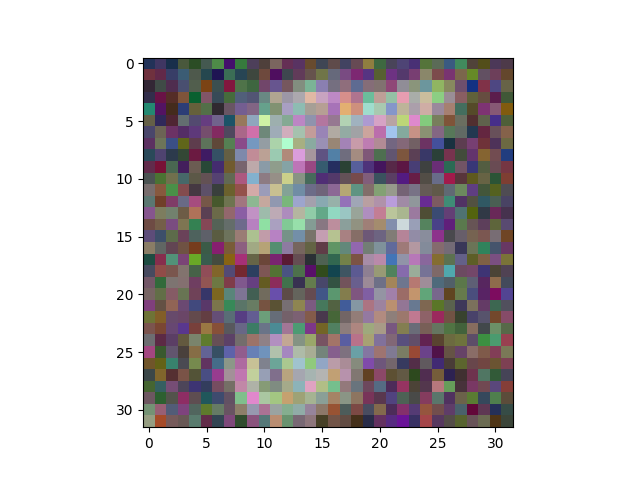}
         \caption{Corrupted image.}
         \label{fig:svhn_corrupt}
     \end{subfigure}
     \begin{subfigure}[b]{0.215\textwidth}
         \centering
         \includegraphics[width=\textwidth]{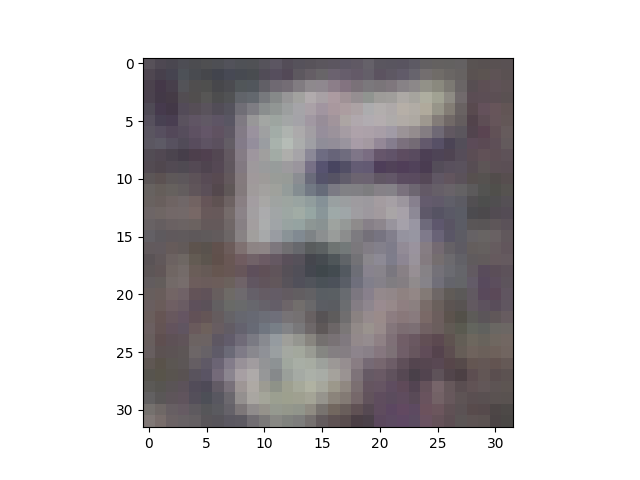}
         \caption{Denoised image.}
         \label{fig:svhn_recon}
     \end{subfigure}
     \vspace{-0.15cm}
        \caption{Example image randomly sampled from a dataset test set (Left), the same image corrupted with noise $\sim \mathcal{N}(0,0.1)$ (Middle), and the Conv-NGC denoising of the corrupted pattern (Right). Top row shows a sample taken from CIFAR-10 while the bottom row shows one taken from SVHN.}
        \label{fig:denoise_samples}
\end{figure}

\noindent
\textbf{Out-of-Distribution Reconstruction:} One interesting property of auto-encoding systems that we are interested in examining is their ability to reconstruct image pattern samples not seen in the original data distribution, particularly samples that come from a much different distribution (violating the typical i.i.d. assumption in statistical learning). To evaluate out-of-distribution (OOD) reconstruction ability, we take the  models trained on the SVHN dataset (the source dataset) and evaluate their ability to reconstruct samples from the distinct CIFAR-10 dataset as well as the larger, more difficult CINIC-10 database (specifically evaluating predictive ability on these test sets). For the results of this OOD task, see Table \ref{table:ood_recon}.

\begin{table}[!htb]
\caption{Both in (a) and (b), mean and standard deviation of SSIM, PSNR, and MSE are reported (over five experimental trials). Lower MSE and higher SSIM and PSNR are better.}
\begin{subtable}{1\columnwidth}
\centering
\begin{tabular}{c| c  | c } 
\hline
 \textbf{CIFAR-10} & \multicolumn{1}{|c}{Reconstruction} & \multicolumn{1}{|c}{Denoising ($\sim \mathcal{N}(0,0.1)$)}\\
 \textbf{Model} & \textbf{PSNR}  & \textbf{PSNR}  \\ 
 \hline\hline
 Conv-AE & $36.7872 \pm 1.0056$ & $24.6527 \pm 0.2001$ \\
 Conv-DAE & $38.6462 \pm 1.2033$  & $\mathbf{36.4346 \pm 0.2021}$ \\
 NGC & $22.8462 \pm 0.0400$  & $19.0410 \pm 0.1203$\\
 Conv-NGC & $\mathbf{41.0912 \pm 0.0234}$  & $25.9004 \pm 0.2004$ \\
\hline
 \textbf{SVHN} & \multicolumn{1}{|c}{Reconstruction} & \multicolumn{1}{|c}{Denoising ($\sim \mathcal{N}(0,0.1)$)}\\
 \textbf{Model} & \textbf{PSNR}  & \textbf{PSNR}  \\ 
 \hline\hline
 Conv-AE & $39.3777 \pm 0.2000$ & $26.2369 \pm 0.3000$ \\
 Conv-DAE & $43.7750 \pm 0.3000$ & $\mathbf{28.6864 \pm 0.2000}$ \\
 NGC & $33.2763 \pm 1.0800 $  & $19.8709 \pm 2.0800$ \\
 Conv-NGC & $\mathbf{45.2305 \pm 0.0800}$  & $28.2513 \pm 0.1600$ \\
 \hline
\end{tabular}
\caption{Analysis of model reconstruction and denoising ability}
\label{table:psnr_analysis}
\end{subtable}

\begin{subtable}{1\columnwidth}
\centering
\begin{tabular}{c| c c c } 
 & \multicolumn{3}{|c}{SVHN to CIFAR-10}  \\
 \textbf{Model} & \textbf{SSIM} & \textbf{PSNR} & \textbf{MSE} \\ 
 \hline\hline
 Conv-AE & $0.69 \pm 0.0010 $ & $24.03 \pm 0.26$ & $387.02 \pm 7.82$ \\
 Conv-DAE & $0.77 \pm 0.0002$ & $29.46 \pm 0.37$ & $176.00 \pm 5.09 $ \\
 Conv-NGC & $\mathbf{0.98 \pm 0.0006}$ & $\mathbf{35.03 \pm 0.30}$ & $\mathbf{26.80 \pm 2.00}$ \\
 \hline
\end{tabular}
\end{subtable}

\begin{subtable}{1\columnwidth}
\centering
\begin{tabular}{c| c c c } 
 & \multicolumn{3}{|c}{SVHN to CINIC-10}\\
 \textbf{Model} & \textbf{SSIM} & \textbf{PSNR} & \textbf{MSE} \\ 
 \hline\hline
 Conv-AE & $0.70 \pm 0.0050$ & $23.74 \pm 0.31$ & $275.09 \pm 8.02$ \\
 Conv-DAE & $0.79 \pm 0.0020$ & $27.10 \pm 0.50$ & $127.90 \pm 4.55$ \\
 Conv-NGC & $\mathbf{0.97 \pm 0.0013}$ & $\mathbf{32.54 \pm 0.20}$ & $\mathbf{47.83 \pm 2.50}$ \\
 \hline
\end{tabular}
\caption{Out-of-distribution reconstruction performance results.}
\label{table:ood_recon}
\vspace{-0.5cm}
\end{subtable}%

\vspace{-0.15cm}

\end{table}

\begin{figure*}[htb!]
     \centering
     \begin{subfigure}[b]{0.75\textwidth}
         \centering
         \includegraphics[width=\textwidth]{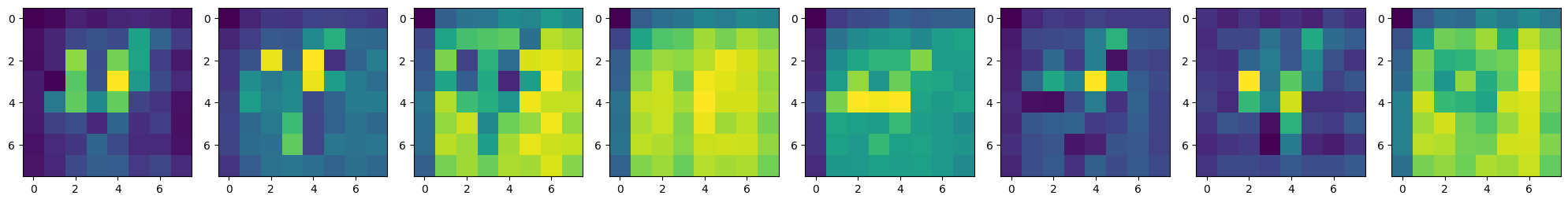}
         \includegraphics[width=\textwidth]{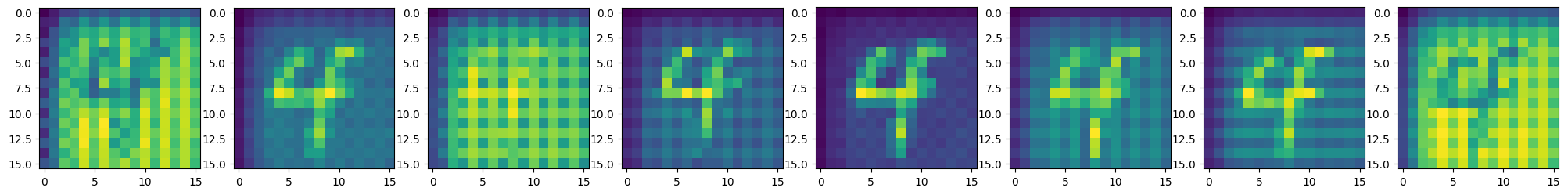}
         \includegraphics[width=0.4\textwidth]{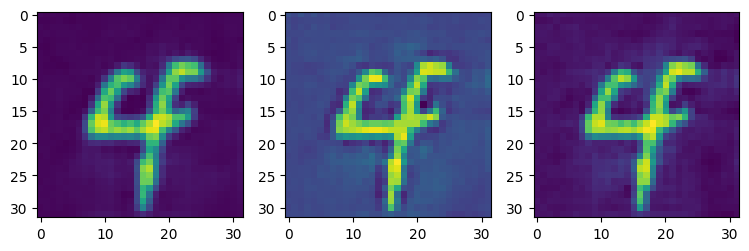}
     \end{subfigure}
     \rulesep
     \begin{subfigure}[b]{0.75\textwidth}
         \centering
         \includegraphics[width=\textwidth]{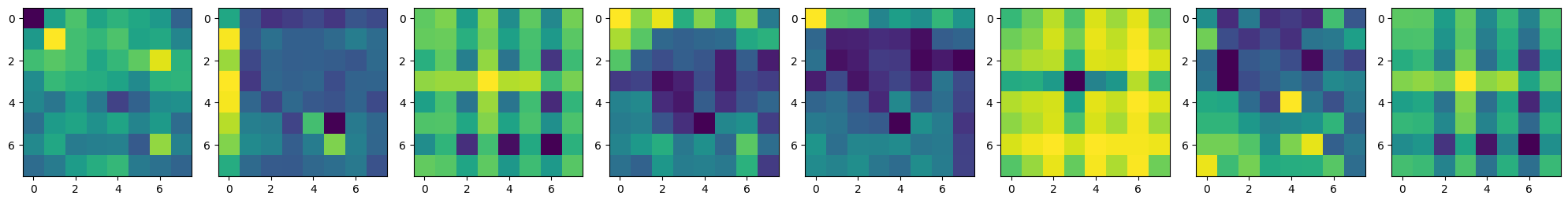}
         \includegraphics[width=\textwidth]{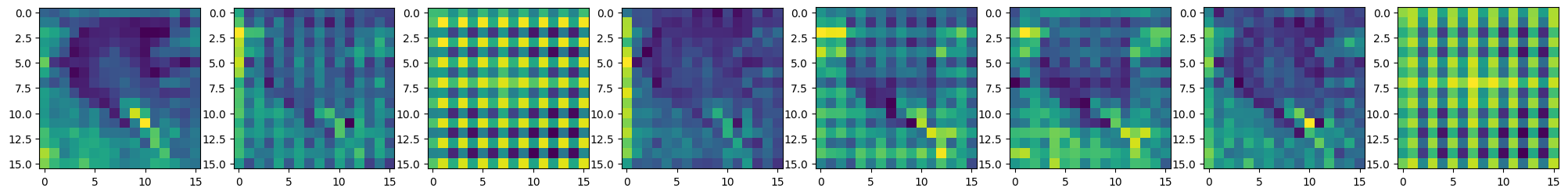}
         \includegraphics[width=0.4\textwidth]{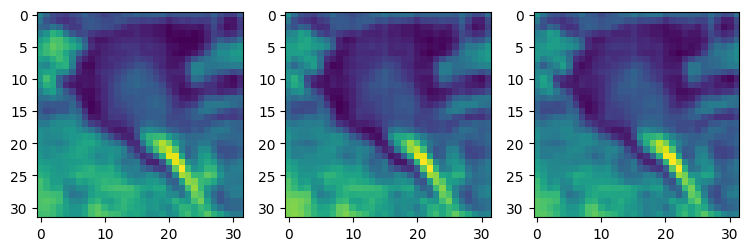}
     \end{subfigure}
     \rulesep
     \begin{subfigure}[b]{0.75\textwidth}
         \centering
         \includegraphics[width=\textwidth]{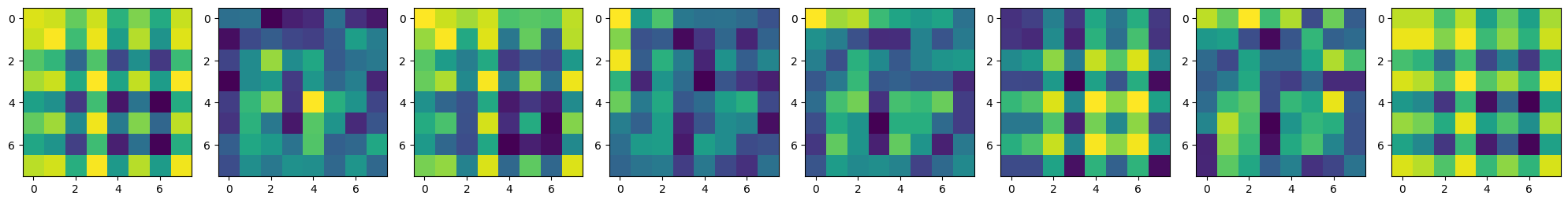}
         \includegraphics[width=\textwidth]{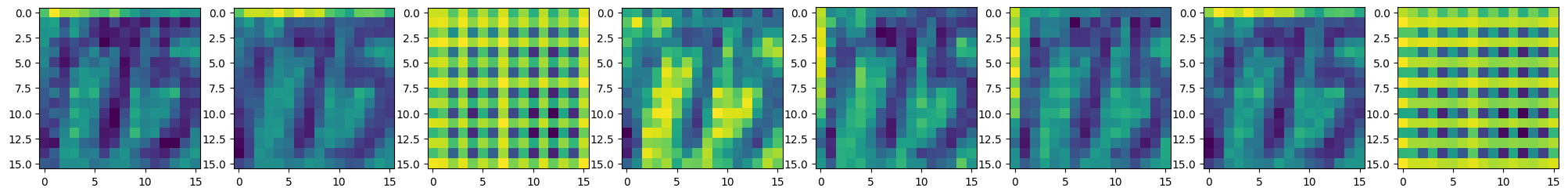}
         \includegraphics[width=0.4\textwidth]{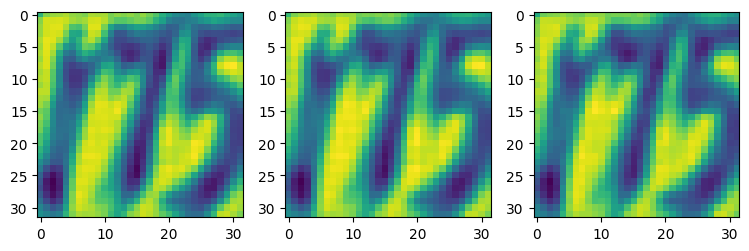}
     \end{subfigure}
        \caption{Features maps for the bottom three layers of a trained Conv-NGC model. (Top 3) Color MNIST feature state maps, (Middle 3) Cifar feature/state maps, and (Bottom 3) SVHN feature/state maps. For each group of three rows, the bottom row shows clamped input, the middle shows the first latent state map layer, and the top shows the second highest latent state map/layer.}
        \label{fig:feature_maps}
\vspace{-0.5cm}
\end{figure*}

\noindent 
\textbf{Discussion: } As seen in our results, Conv-NGC outperforms NGC (or PCN), as expected, and exhibits competitive behavior with the backprop-based autoencoder models. Notably, in terms of reconstruction, Conv-NGC even offers improved SSIM and PSNR, which is likely the result of its ability to learn a reconstruction process (over a $T$-length window of time). With respect to image denoising, we observe that the Conv-NGC models works well, outperforming both the AE and NGC models specifically on the harder natural image datasets (CIFAR-10 and SVHN), but under-performs the DAE. The DAE, however, was trained directly for the task of denoising (with noise injected to its input nodes), so it makes sense that it would outperform models that were not tuned to the task (and thus serves well as a soft upper bound on performance). See Figure \ref{fig:denoise_samples} for visual examples of image samples that the Conv-NGC model denoised (for CIFAR-10 and SVHN).

Note that, for the table metrics reported, pixel reconstruction/model output values were re-scaled to be between $[0,255]$, which better highlighted the gap between results obtained for Conv-NGC and other baseline models. More importantly, Conv-NGC exhibits a low variance and better visual reconstruction/modeling comparatively as indicated by SSIM, which is a metric that more closely relates to / correlates with human perception. 
To probe the knowledge acquired in the Conv-NGC's distributed representations, we examined the learned latent state feature maps of trained models on sampled test images, taken from each of the data benchmarks. Qualitatively, we observed that Conv-NGC appears to learn a (noisy) form of the image pyramid \cite{adelson1984pyramid} within its latent activities (see Figure \ref{fig:feature_maps} for visual samples of this result). 
Notably, in Conv-NGC, the sensory image appears in some of the internal feature state maps, but is a down-sampled, decreased resolution form of itself (much akin to the repeated process of smoothing/sub-sampling that results in lower spatial density images the higher up one goes within the image pyramid). This result complements early work that highlighted the notion that the brain processes visual information at different resolutions \cite{campbell1968application}. This biological multi-resolution analysis of the world allows the brain to extract useful information from complex input patterns and it appears that Conv-NGC implicitly learns to conduct a similar type of analysis on the patterns contained in the natural image benchmarks. 

Surprisingly, with respect to the OOD experiments, Conv-NGC outperforms all of the baseline models  (including the Conv-DAE), further corroborating the in-dataset reconstruction result(s) that Conv-NGC appears to (meta-)learn a type of reconstruction process that works well on unseen, out-of-distribution natural image patterns. This desirable improvement is indicated by the higher trial-averaged (OOD) PSNR as well as the (OOD) SSIM score on both problem settings: 
1) SVHN-to-CIFAR-10 reconstruction, and 
2) SVHN-to-CINIC-10 reconstruction. 
What is most impressive is that the Conv-NGC model was trained on natural images (in SVHN) that did not contain any of the objects found within either CIFAR-10 and CINIC-10 and it was still able to reconstruct with top performance. This result offers a promising future direction worth exploring for Conv-NGC/NGC predictive coding systems in general -- their ability to generalize to unseen patterns that violate the assumption that they were generated by a distribution similar to that of the training data.

\section{Conclusion}
\label{sec:conclusion}

In this work, we proposed convolutional neural generative coding (Conv-NGC), a generalization of a computational predictive coding framework to the case of natural images. Our experiments on three benchmark datasets, i.e., Color-MNIST, CIFAR-10, and SVHN, demonstrate that Conv-NGC is competitive with powerful backprop-based convolutional autoencoding models with respect to both pattern reconstruction and image denoising and notably outperforms all of them in the context of out-of-distribution reconstruction. Our results mark an important step towards crafting more robust, general brain-inspired neural architectures and learning processes capable of handling complex machine learning tasks.

\bibliographystyle{acm}
\bibliography{ref}

\newpage
\section*{Appendix-A}
In this section we compare and contrast Conv-NGC with other inference and credit assignment approaches such as predictive coding/processing and local learning.

\subsection*{Related Work}

\textbf{Predictive Coding: }
The motivation behind our model's design and learning procedure is grounded in predictive coding principles~\cite{rainer1999prospective}. According to predictive coding theory~\cite{spratling2010predictive,brodski2015faces,rao1999predictive,friston2010free}, at each level of a cognitive process, the brain continuously generates and updates beliefs about what information its internal model of the world should be receiving from the level above it. These beliefs get translated into predictions, which are the best possible explanations of what is perceived/observed, such that the system's overall (sensory) experience makes sense~\cite{panichello2013predictive}. Sparse predictive coding models~\cite{olshausen1997sparse,rao1999predictive} embody the hypothesis that the brain is a type of directed generative model, combining top-down prediction and bottom-up signals/messages in order to facilitate dynamic adaptation to to its environment~\cite{rauss2013predictive}. This forms a type of iterative inference that progressively corrects latent distributed representations. Furthermore, when nesting the ideas of predictive coding within the Kalman Filter framework~\cite{rao1997dynamic}, one can create dynamic models that handle time-varying data. Many variations and implementations of predictive coding have been developed~\cite{chalasani2013deep,chalasani2015contextcdn,lotter2016deep,santana2017exploiting,hwang2018robotPC,song2018fastinfpc, ororbia&mali2019lifelong,ororbia2018continual,salvatori2022reverse}. Some of the more recent ones merge it with back-propagation of errors as a subsequent fine-tuning step and speed up training~\cite{mittal2020learning,oord2018representation}. Nevertheless, most prior work on predictive coding networks (with notable exceptions such as \cite{spratling2017hierarchical}), in general, focuses on neural coding blocks implemented as fully connected layers \cite{whittington2017approximation,salvatori2021associative,ororbia2022neural,gklezakos2022active} while others that do utilize convolution adapt/modify backprop-based networks to utilize principles of predictive coding \cite{lotter2016deep,han2018deep,wen2018deep,choksi2021predify}. In this work, we are the first to generalize predictive coding fully to the case of (de)convolution (and residual blocks as shown in the Appendix) for visual data processing, specifically facilitating the design of causal-like, decoder-only generative/reconstruction models of natural images.

\textbf{Local Learning: }   
One key concept behind predictive coding is that each neural architecture layer would need an error feedback mechanism to communicate its needs, i.e., activity mismatch signals transmitted to relevant nearby layers. Suppose the feedback signals come closer to the layers themselves. In this scenario, error connections can share information to the right units such that the discrepancy between learned representations and target representations is minimum. Such types of local updates are referred to as local learning approaches~\cite{ororbia2018biologically,lee2015difference,nokland2016direct}, which offer a potential replacement for back-propagation when training artificial neural networks. Empirically, updates from a more localized approach can lead to better generalization performance~\cite{lee2015difference,ororbia2018biologically}, even with temporal data \cite{mali2021investigating} and discrete signals \cite{mali2021empirical}. Historically, there have been many efforts at crafting algorithms that learn locally, including those based on contrastive Hebbian learning \cite{movellan1991contrastive}, contrastive divergence for learning harmoniums \cite{hinton2002training}, the wake-sleep algorithm for adapting Helmholtz machines \cite{hinton1995wake}, and, more recently, equilibrium propagation \cite{scellier2017equilibrium}.  
Other recent efforts, which more directly integrate local learning into the deep learning pipeline/modeling framework, include kickback~\cite{balduzzi2015kickback} and decoupled neural interfaces~\cite{jaderberg2016decoupled}.

\section*{Appendix-B}

In this section we provide dataset statistics and also provide information concerning hyper-parameters and best settings for our models including baseline models. We later mathematically define the metrics used to compare performance for each model on various benchmarks as shown in main paper. Finally, we provide qualitative analysis by visualizing the feature state maps learned by our model by conducting a feature map analysis by extracting intermediate filters from each synapse. 

\subsection*{Baseline Models and Setup}

\noindent
\textbf{Convolutional Autoencoder:} 
As our first baseline, we designed a convolution auto-encoder (Conv-AE) to compare to our Conv-NGC decoder-only model. The Conv-AE consists of five blocks in encoder, where one block is compromised of a convolution layer (with $32$ output feature maps,  $3x3$ kernel, ``SAME'' padding, and a stride of $2$) followed by a relu activation, batch normalization and finally a max-pooling layer (also with a stride of $2$). The bottleneck layer is then passed through multiple layers of deconvolution in order to reconstruct the input. The AE models were trained using Adam with a step size of $2e-5$ (mini-batches used to estimate gradients were of size $128$) with weights initialized using glorot initialization. The output layer of the decoder uses a logistic sigmoid activation. Conv-AEs have total of $48,393$ trainable parameters in order to learn the distribution of each natural image benchmark.

\noindent
\textbf{Convolutional Denoising Autoencoder:}
Beyond the Conv-AE, we designed a convolutional denoising autoencoder (Conv-DAE), which follows the same design principle as the Conv-AE but with the key difference that we inject noise into the input data (to make model robust against noise). This noise injection process helps the DAE to better reconstruct inputs by averaging over pixels to create a smoother output. Another notable alteration made to the Conv-DAE was that the selu activation function was used instead of the relu. Based on preliminary experimentation, we observed that replacing relu with selu resulted in overall better reconstruction for Conv-DAEs. Conv-DAEs have a total of $48,393$ trainable parameters in order to learn the distribution of each benchmark.

\noindent
\textbf{Neural Generative Coding:} We also experimented with the fully-connected form of neural generative coding (NGC), which, when the error feedback synaptic matrices are set to be equal to the transpose of the forward generative synaptic matrices and the derivative of the activation is used in the latent state update, we recover the unsupervised form of the model in \cite{salvatori2021associative} (also referred to as the hierarchical predictive coding network, or PCN). For all datasets studied, the NGC model we crafted contained five state layers (the bottom layer was sized to be the same shape as the input image) where each layer contained $200$ neurons and employed a leaky rectifier activation for $\phi^\ell(\mathbf{v})$ with generative synapses initialized from a zero-mean Gaussian distribution and standard deviation of $0.1$ (biases were initialized to zero vectors). The output prediction activation were set to be the identity, i.e., $g^\ell(\mathbf{v}) = \mathbf{v}$, except for the bottom layer ($g^0(\mathbf{v})$), which was set to be the logistic sigmoid. The NGC model's stimulus window was set to be of length $T=60$ and a state update rate of $\beta = 0.1$ was used. Latent state vector values were sampled from a Gaussian distribution $\mathcal{N}(0.5,0.05)$ (meaning at the first step of the stimulus window, they were non-zero). 
Synaptic parameters were adjusted using the Adam adaptive learning rate with step size $0.001$  with mini-batches of size $500$. In accordance with \cite{ororbia2022neural} and with proper sparse coding practice, we constrained the L2 norms of the columns of the synaptic matrices to have values no greater than one. 
Note that, for predicting $32\times32\times3$ color images, our NGC models utilized $738,072$ total free parameters to learn the distribution of each benchmark (considerably more synaptic parameters than the convolutional autoencoder models).

\noindent
\textbf{Convolutional Neural Generative Coding:} For convolutional neural generative coding (Conv-NGC), for all datasets, we designed all models to contain five layers of state maps, where each layer/block leveraged deconvolution with a stride set to $s =2$ (kernel size was $3\times3$ using ``SAME'' padding), meaning that each map in the top-most latent state layer was of size $2\times2$ neurons and increased in size until reaching a size of $32\times32$ at the bottom layer (where each map in the bottom-most layer, which is normally be clamped to an input image color channel, would also be of size $32\times32$). The channel configuration of our Conv-NGC decoder-only model is $\{10,15,20,25,3\}$ (going from left to right gives the number of channels starting from the top-most latent state layer, where the rightmost value should be equal to the number of color channels that define a natural image). 
The state activation function $\phi^\ell()$ was set to be the leaky rectifier while the local prediction activation $g^\ell()$ was set to be the identity (for all layers including the bottom-most layer). Synapses were initialized from a zero-mean Gaussian distribution with a standard deviation of $0.1$ (bias maps were initialized to zero values). The Conv-NGC model's stimulus window was set to a length of $T=60$ and a state map update rate of $\beta = 0.1$ was used. The top-most latent state maps were sampled from a Gaussian distribution $\mathcal{N}(0.5,0.05)$ while the rest were initialized from the top-down ancestral sampling pass (see the algorithm in the Appendix for details). 
Synaptic matrices were adjusted using Adam with a step size of $0.001$ using mini-batches of size $500$. 
In contrast to the NGC models, Conv-NGC utilized a total of $9,225$ free parameters in order to learn the distribution of each benchmark (note that if separate error filters were used, this would only double the parameter count to a mere $18,450$ synapses), considerably fewer parameters than the NGC and autoencoder models. Finally, note that in the case of denoising, the bottom-most units/state maps of both NGC and Conv-NGC were only clamped to the input image(s) at the first time step $t=1$ of the stimulus and allowed to evolve according to the dynamics of Equation \ref{eq:correction}.

\subsection*{Measured Metrics}

\noindent
\textbf{Reconstruction and Denoising: }
In this set of tasks, we investigate each model's ability to reconstruct the given input image patterns from each dataset as well as to recover the original input image values under corruption noise. 
For the task of reconstruction, we measure the mean squared error (MSE) for each input between the model's predicted/output values $\mathbf{\hat{x}}_i$ and the original image $\mathbf{x}_i$ as follows:
\begin{align}
    \text{MSE}(\mathbf{x_i},\mathbf{\hat{x_i}}) = \frac{1}{I} \sum^I_{i=1} \big( \frac{1}{C} \sum^C_{c=1} \big( \frac{1}{N*M} \sum^N_{n=1} \sum^M_{m=1} \big( \mathbf{\hat{x}}_{nmci} - \mathbf{x}_{nmci} \big)^2 \big) \big)
\end{align}
which is calculated per channel $c \in C$ and then averaged over all channels and images in the test set. To better probe the image quality of the reconstructed outputs, we evaluate the structural similarity index measure (SSIM):
\begin{align}
    \text{SSIM}(\mathbf{x}_i, \mathbf{\hat{x}}_i) = \frac{ (2\mu_{x_i}\mu_{\hat{x}_i} + (0.01 Q)^2) (2\sigma_{x_i \hat{x}_i} + (0.03 Q)^2) }{(\mu_{x_i}^2 + mu_{\hat{x}_i}^2 + (0.01 Q)^2) (\sigma_{x_i}^2 + \sigma_{\hat{x}_i}^2 + (0.03 Q)^2)}
\end{align}
where $\mu_{x_i}$ and $\mu_{\hat{x}_i}$ are the pixel sample means for the original image $\mathbf{x}_i$ and reconstruction $\mathbf{\hat{x}}_i$, respectively, and $\sigma_{x_i}$ and $\sigma_{\hat{x}_i}$ are the pixel standard deviations ($\sigma_{x_i \hat{x}_i}^2$ is the covariance between $\mathbf{x}_i$ and $\mathbf{\hat{x}}_i$). $Q$ is the dynamic range of the pixels, i.e., $Q = 2^{\text{bpc}}-1$ (where $\text{bpc}$ is bits-per-pixel).

Finally, for the more complex natural images in SVHN and CIFAR-10, we further measure peak signal-to-noise ratio (PSNR) formally defined as follows:
\begin{align}
   \text{PSNR}(\mathbf{x}_i, \mathbf{\hat{x}}_i) = \frac{1}{I} \sum^I_{i=1} \big( \frac{1}{C} \sum^C_{c=1} \big( 20 \times \log_{10}  \big( 255.0/ \notag \\ \big( \sqrt{   \frac{1}{N*M} \sum^N_{n=1} \sum^M_{m=1} \big( \mathbf{\hat{x}}_{nmci} - \mathbf{x}_{nmci} \big)^2} \big) \big) \big) \big) .
\end{align}

\subsection*{Dataset Descriptions}
\label{sec:data}

\subsubsection*{Datasets}

We next describe the data benchmarks that were used to evaluate Conv-NGC and the relevant baseline systems. The first three, Color MNIST, CIFAR-10, and SVHN, were utilized in both training and testing/evaluation contexts while the fourth one, CINIC-10, was only used (as a test set for the out-of-distribution reconstruction experiments).

\noindent
\textbf{Color-MNIST:} Color-MNIST is the colored version of the MNIST dataset with $60,000$ training and $10,000$ test samples. We resize Color-MNIST images to $32 \times 32$ using linear interpolation. In the original MNIST, the images are grayscale; in the colored form of MNIST, each image is either red or green. This database is designed to correlate strongly (though spuriously) with the class label. Thus, by construction, the label is more strongly correlated with the color than with the digit, so any algorithm purely minimizing training error will likely exploit the color. Such approaches will fail at test time because the direction of the correlation is reversed in the test environment, which makes this dataset a challenging benchmark.

\noindent
\textbf{The CIFAR-10 Dataset:} 
The CIFAR-10 dataset has $50,000$ training and $10,000$ test images, across $10$ categories including those depicting animals and vehicles.  Images are of size $32\times32$ pixels. $5,000$ training samples were set aside to measure validation metrics and tune hyper-parameters. 

\noindent
\textbf{The Street View House Numbers (SVHN) Dataset:} The SVHN dataset has approximately $75,000$ training and $23,000$ test images. All images are of size $32\times 32 \times 3$ pixels. We randomly select $5,000$ samples from all $10$ classes to create validation set to perform hyper-parameter optimization.

\noindent
\textbf{The CINIC-10 Dataset:} The CINIC-10 dataset is a more challenging alternative to CIFAR-10 and contains a total of $270,000$ natural color images, nearly $4.5$ times that of the size of CIFAR-10. The CINIC database is ultimately built from two data sources, i.e., ImageNet and CIFAR-10. The data is further split into three equal subsets, i.e., a training, validation, and testing partition, where each contains $90,000$ images. (As noted earlier, we only use its test-set for the out-of-distribution prediction experiments later.)

\begin{algorithm*}[!t]
    \caption{Conv-NGC state prediction and correction.}
    \label{algo:inference}
    \begin{algorithmic}[1]
    \State {\bfseries Input:} Data sample $(\mathbf{x},\mathbf{y})$; model parameters $\Theta$; constants $\beta$, $\gamma$, and $T$
    \State {\bfseries Output:} Corrected states/maps $\{\mathbf{z}^0, \mathbf{z}^1, \mathbf{z}^2, \cdots, \mathbf{z}^L \}$; error units/groups $\{\mathbf{e}^0, \mathbf{e}^1, \cdots, \mathbf{e}^{L} \}$ 

    \LineComment{Initialize system with an ancestral (top-down) projection pass}
    \State $\mathbf{z}^{L}_c \sim \mathcal{N}(\mu_z,\sigma_z), \; c = 1,...,C_L \; \text{and} \; \mathbf{e}^{L}_c = \mathbf{0}, \; c = 1,...,C_L$, \; $\mathbf{z}^0 = \mathbf{x}$
    \For{layer $\ell = L-1$ to $0$}
        \For{filter $c = 1$ to $C_\ell$} \Comment For each output filter in layer $\ell$
            \State $\mathbf{z}^{\ell}_c = \mathbf{0}$
            \For{filter $i = 1$ to $C_{\ell+1}$} \Comment For each input filter in layer $\ell+1$
                \State $\mathbf{\bar{z}}^{\ell}_c \leftarrow \mathbf{\bar{z}}^{\ell}_c + \Big( \mathbf{W}^{(\ell+1)}_{ic} \circlearrowleft_s \phi^{\ell+1}( \mathbf{z}^{(\ell+1)}_i) \Big)$ \Comment Partially compose state map of layer below
            \EndFor
            \State $\mathbf{z}^{\ell}_c = \mathbf{\bar{z}}^{\ell}_c = g^{\ell+1}(\mathbf{\bar{z}}^{\ell}_c + \mathbf{b}^{\ell+1}_c)$
        \EndFor
    \EndFor
    \LineComment{Simulate iterative processing over stimulus window of length $T$}
    \For{$t = 1$ to $T$}
        \For{layer $\ell = L-1$ to $0$}
            \LineComment{Run layer-wise predictors (per state feature map)} 
            \State $\mathbf{\bar{z}}^{\ell}_c = \mathbf{0}, \; c = 1,...,C_\ell$
            \For{filter $c = 1$ to $C_\ell$} \Comment For each filter in layer $\ell$
                \For{channel $i = 1$ to $C_{\ell+1}$} \Comment For each input channel in layer $\ell$
                    \State $\mathbf{\bar{z}}^{\ell}_c \leftarrow \mathbf{\bar{z}}^{\ell}_c + \Big( \mathbf{W}^{(\ell+1)}_{ic} \circlearrowleft_s \phi^{\ell+1}( \mathbf{z}^{(\ell+1)}_i) \Big)$ \Comment Update output channel/prediction $c$ of layer $\ell$
                \EndFor
                \State $\mathbf{\bar{z}}^{\ell}_c = g^{\ell+1}(\mathbf{\bar{z}}^{\ell}_c + \mathbf{b}^{\ell+1}_c), \; \mathbf{e}^\ell_c = (\mathbf{z}^\ell_c - \mathbf{\bar{z}}^\ell_c)$ \Comment Calculate layer-wise error units (per feature map)
            \EndFor
            \LineComment{Correct internal states (per state feature map)}
            \State $ \mathbf{d}^{\ell+1}_c = \mathbf{0}, \; c = 1,...,C_{\ell+1}$
            \For{channel $i = 1$ to $C_{\ell+1}$} \Comment For each input channel in $\ell+1$, compute (total) perturbation
                \For{channel $c = 1$ to $C_{\ell}$} \Comment For each output (error) channel in layer $\ell$
                    \State $ \mathbf{d}^{\ell+1}_i \leftarrow \mathbf{d}^{\ell+1}_i + (\mathbf{E}^{\ell+1}_{ci} *_s \mathbf{e}^{\ell}_c)$
                \EndFor
                \State $ \mathbf{d}^{\ell+1}_i = \mathbf{d}^{\ell+1}_i - \mathbf{e}^{\ell+1}_i$
                \; and \; $\mathbf{z}^{\ell+1}_i \leftarrow ( \mathbf{z}^{\ell+1}_i + \beta \mathbf{d}^{\ell+1}_i - \gamma \mathbf{z}^{\ell+1}_i ) $
            \EndFor
    	\EndFor
    \EndFor
    \State \textbf{Return:} $\{\mathbf{z}^0, \mathbf{z}^1, \mathbf{z}^2, \cdots, \mathbf{z}^L \}$, $\{\mathbf{e}^0, \mathbf{e}^1, \cdots, \mathbf{e}^{L-1} \}$ \Comment Output system statistics
    \end{algorithmic}
\end{algorithm*}

\begin{algorithm*}[!t]
    \caption{Conv-NGC synaptic adjustment (given statistics from Algorithm \ref{algo:inference}).}
    \label{algo:update}
    \begin{algorithmic}[1]
    \State {\bfseries Input:} State maps $\{\mathbf{z}^0, \mathbf{z}^1, \mathbf{z}^2, \cdots, \mathbf{z}^L \}$; error maps $\{\mathbf{e}^1, \mathbf{e}^2, \cdots, \mathbf{e}^L \}$; \text{model parameters} $\Theta$;  
    \State \hspace{0.9cm} Constants $\lambda,\;\alpha,\;\gamma,\;\epsilon$
    \State {\bfseries Output:} Updated $\Theta$
    \For{layer $\ell = 1$ to $L$}
        \LineComment{Calculate synaptic adjustments/displacements}
        \For{layer $c = 1$ to $C_\ell$} \Comment For each filter in layer $\ell$
            \State $\Delta \mathbf{W}^\ell_{ic} = \mathbf{0}$, \; $\Delta \mathbf{E}^\ell_{ci} = \mathbf{0}$
            \For{layer $i = 1$ to $C_{\ell-1}$} \Comment For each input (error) channel of layer $\ell-1$
                \State $\Delta \mathbf{W}^\ell_{ic} \leftarrow \Delta \mathbf{W}^\ell_{ic} + \mathbf{e}^{\ell-1}_c *_{1} \text{Dilate}\Big( \big( \phi^\ell(\mathbf{z}^{\ell}_i) \big)^T, s \Big)$
                \State $\Delta \mathbf{E}^\ell_{ci} \leftarrow \Delta \mathbf{E}^\ell_{ci} + \lambda \Big(  \text{Dilate}\Big( \big( \phi^\ell(\mathbf{z}^{\ell}_i) \big)^T, s \Big) *_1 \mathbf{e}^{\ell-1}_c \Big)$
            \EndFor
            \LineComment{Update current filter synapses -- could alternatively use Adam \cite{kingma2014adam}} 
            \State $ \mathbf{W}^\ell_{ic} \leftarrow  \mathbf{W}^\ell_{ic} - \alpha (\Delta  \mathbf{W}^\ell_{ic} / (||\Delta  \mathbf{W}^\ell_{ic}||_2 + \epsilon) )$ \; and \; $ \mathbf{E}^\ell_{ci} \leftarrow  \mathbf{E}^\ell_{ci} - \gamma (\Delta  \mathbf{E}^\ell_{ci} / (||\Delta  \mathbf{E}^\ell_{ci}||_2 + \epsilon) )$
        \EndFor
    \EndFor
    \State \textbf{Return:} $\Theta = \big\{ \{ \{\mathbf{W}^\ell_{ic}, \mathbf{E}^\ell_{ci} \}^{C_\ell}_{i=1} \}^{C_{\ell-1}}_{c=1} \big\}^L_{\ell=1}$ \Comment Output newly adjusted synaptic efficacies
    \end{algorithmic}
\end{algorithm*}

\section*{Appendix-C}
We provide an algorithmic specification of Conv-NGC's core processes (inference/synaptic adjustment) as well as details on how to implement Conv-NGC in the form of residual block structures instead of the (de)convolutional form in the main paper. Furthermore, Table \ref{table:definitions} contains definitions of key acronyms/variables/operators used in the main paper.

\noindent 
\textbf{Conv-NGC Algorithmic Specification: }
In Algorithms \ref{algo:inference} and \ref{algo:update}, we provide the pseudocode detailing the key processes of iterative inference as well as synaptic adjustment for the Conv-NGC models we experimented with in the main paper. Note that the notation/operators/symbols used are the same as those utilized in the main paper.

\subsection*{Inference: Predicting and Correcting Neural Residual Blocks} 
\label{subsec:residual_prediction}

\textbf{State Block Prediction:} 
Extending the prediction operation of Equation \ref{eqn:prediction} to make use of stacks of operations, such as those that compose a residual block, is simple -- the $i$th state feature map at $\ell$, i.e., $\text{ReLU}(\mathbf{z}^\ell_i)$, is run through a series of convolution and pooling operations instead of just the single convolutional layer originally depicted in order to obtain a prediction of the $i$th state feature map at layer $\ell-1$. For example, the residual block $R_s()$ we designed and employed in this paper entailed the following chain of operations:
\begin{align} 
    \mathbf{h}^{\ell,4}_j &= \Big( \sum^{C_{\ell}}_i \mathbf{W}^{\ell,4}_{ij} \circlearrowleft_s \text{ReLU}( \mathbf{z}^{\ell}_i ) \Big) \nonumber \\
    \mathbf{h}^{\ell,3}_j &= \text{ReLU}\Big( \text{Pool}\Big( \sum^{C^4_{\ell}}_i \mathbf{W}^{\ell,3}_{ij} \circlearrowleft_s \mathbf{h}^{\ell,4}_i \Big) \Big), \; 
    \mathbf{h}^{\ell,2}_j = \text{Pool}\Big( \sum^{C^3_{\ell}}_i \mathbf{W}^{\ell,2}_{ij} \circlearrowleft_s \mathbf{h}^{\ell,3}_i \Big) \nonumber \\
    \mathbf{\bar{z}}^{\ell-1}_j &= \mathbf{h}^{\ell,1}_j = \text{ReLU}\Big( \sum^{C^2_{\ell}}_i \mathbf{W}^{\ell,1}_{ij} \circlearrowleft_s \mathbf{h}^{\ell,2}_i \Big) + \mathbf{z}^\ell_i
    \label{eqn:residual_prediction}
\end{align}
where $\text{Pool}(\mathbf{v})$ denotes the max-pooling operation. Notice that $\mathbf{h}^{\ell,q}$ represents an ``intermediate'' layer within the ConvNCNet, or rather, an activity that exists within the residual block but does not count as an actual state. Specifically, $s$ is the skip length of the residual block (in Equation \ref{eqn:residual_prediction} $s = 5$) and $\mathbf{h}^{\ell,q}_i$, where $q = \{ s-1,s-2,\cdots,1 \}$, which denotes the $q$th intermediate layer within the residual that relates $\mathbf{z}^\ell_i$ and $\mathbf{z}^{\ell-1}_i$. 
$W^{\ell,q}_{ij}$ 
denotes the $i$th learnable filter associated with the $q$th intermediate layer calculation within the residual block. 
$C^q_\ell$ is the number of channels that make up an intermediate residual block layer $\mathbf{h}^q_\ell$.

\begin{table*}[!t]
\caption{Table of key symbol/operator/abbreviation definitions.}
\label{table:definitions}
\vspace{-0.3cm}
\begin{center}
\begin{tabular}{||c c||} 
 \hline
 \textbf{Item} & \textbf{Explanation} \\ [0.5ex] 
 \hline\hline
 Conv-NGC & Convolution neural generative coding (model) \\ 
 \hline
 $*_s$ & strided convolution where $s$ is the stride argument\\
 \hline
  $\circlearrowleft_s$ & denotes deconvolution (or transposed convolution) with a stride of $s$ \\
  \hline
 $\cdot$ & Matrix/vector multiplication \\
 \hline
 $\odot$ & Hadamard product (element-wise multiplication) \\
 \hline
 $()^T$  & denotes the transpose operation \\
 \hline
 $\text{Flatten}(\mathbf{z})$ & input tensor $\mathbf{z}$ is converted to a column vector \\
 \hline
 $\text{UnFlatten}(\mathbf{z})$ & is inverse of $\text{Flatten}(\mathbf{z})$ \\
 \hline 
 $\text{Dilate}(\mathbf{v}, s)$ & represents a dilation function controlled by the dilation (integer) size $s$ \\
\hline 
$\mathbf{W}_{ijkl}$ & represents a 4-dimensional tensor or synaptic weights \\
\hline
$\mathbf{E}_{ijkl}$ & represents a 4-dimensional tensor or synaptic error weights \\
\hline
$L$ & denotes set of predictive layers \\
\hline
$\mathbf{z}^{\ell}$ & denotes state representation at layer $\ell$ \\
\hline
$\mathbf{\bar{z}}^{\ell-1}$ & denotes state prediction at layer $\ell$ \\
\hline
$\mathbf{e}^{\ell-1}$ & denotes the error/mismatch signal at layer $\ell-1$ \\
\hline
$\phi^\ell()$ & denotes the activation/transfer nonlinear activation applied to the state tensor \\
\hline
$g^\ell()$ & denotes the activation/transfer nonlinear activation applied to the state tensor prediction \\
\hline
$\leftarrow$ & denotes a variable override and the modulation/adjustment factor \\
\hline 
$\beta$ & denotes constant value, which controls the state correction rate \\
\hline 
$\gamma$ & denotes the strength factor to control leak variable \\
\hline
$T$ & Total number of iterations for iterative inference in Conv-NGC \\
\hline
$\lambda$ & denotes the modulation factor to control the evolution of the error filters \\
\hline
 
\end{tabular}
\end{center}
\end{table*}

\textbf{State Block Correction:} 
State map correction in a residual blocks proceeds a bit differently from Equation \ref{eq:correction} in the main paper, requiring further modification of the perturbation vector term  $\mathbf{d}^\ell_i$. Specifically, for the residual block we presented in Equation \ref{eqn:residual_prediction} above, this correction would become:
\begin{align}
    \mathbf{d}^\ell_i &= -\mathbf{e}^\ell_i + (\sum^{C^3_\ell}_j \mathbf{E}^{\ell,4}_{ji}  \mathbf{d}^{\ell,3}_j ), \\
    \mathbf{d}^{\ell,3}_j &= \sum^{C^2_\ell}_j \mathbf{E}^{\ell,3}_{ji} \mathbf{d}^{\ell,2}_j, \\
    \mathbf{d}^{\ell,2}_j &= \sum^{C^1_\ell}_j \mathbf{E}^{\ell,2}_{ji} \mathbf{d}^{\ell,1}_j, \\
    \mathbf{d}^{\ell,1}_j &= \sum^{C_{\ell-1}}_j \mathbf{E}^{\ell,1}_{ji} \mathbf{e}^{\ell-1}_j \label{eqn:residual_correction}
\end{align}
where we observe that the learnable error synapses of the original LRA algorithm have been adapted to create a multi-step transmission pathway, i.e., a proxy teaching signal (instead of a layer-wise error neuron vector) is created for each intermediate layer, yielding a sort of hybrid between recursive LRA~\cite{ororbia2020large} and feedback alignment \cite{lillicrap2014random}. One could also view Equation \ref{eqn:residual_correction} as one single error transmission pathway that is decomposed into a series of intermediate perturbation vectors (this view will be useful for crafting a fast update rule in the next sub-section). One key advantage of LRA is its flexibility in designing information transmission pathways that do not necessarily need to strictly mirror the flow of the forward transmission pathway~\cite{ororbia2020large}. In Equation \ref{eqn:residual_correction}, this flexibility comes into play given that the error transmission pathway skips over non-learnable operations such pooling and activation functions, quickly facilitating the calculation of the perturbation needed to correct feature map state $\mathbf{z}^\ell_i$. 

\subsection*{Training: Updating Model Parameters}
\label{subsec:residual_update}

\textbf{Residual Block Update:} To train the residual block synapses, we would adapted the LRA algorithm to train the intermediate transformations within the block. Specifically, we take further advantage of the multi-step error transmission pathway presented in Equation \ref{eqn:residual_correction} to craft a simple update rule (specifically, using recursive-LRA~\cite{ororbia2020large} offers the greatest amount of flexibility in the design of error transmission pathways). 
As a result, the rule for adapting the parameters of a residual block is:
\begin{align}
    \Delta \mathbf{W}^{\ell,q}_{ij} &= \mathbf{d}^{\ell,q-1}_j *_{1} \text{Dilate}\Big( \big( \phi^\ell(\mathbf{h}^{\ell,q}_i) \big)^T, s \Big), \; \Delta \mathbf{E}^\ell_{ji} =  \lambda \Big(  (\Delta \mathbf{W}^{\ell,q}_{ij})^T \Big), \\
    &\text{where } \mathbf{d}^{\ell,0}_j = \mathbf{e}^{\ell-1}_j, \; \text{and } \; \mathbf{h}^{\ell,s}_i = \mathbf{z}^\ell_i  \mbox{.} \nonumber
\end{align} 
The key difference between the neural coding and residual block updates is the use of $\mathbf{d^{\ell,q}_i}$ for the weight/filter update. 
As shown in Equation \ref{eqn:residual_correction},  $\mathbf{d}^{\ell,q}_i$ is computed using errors from the layer below it and simply represents a perturbation that could be applied to its respective layer/transformation to help it better predict the downstream target $\mathbf{z}^{\ell-1}_i$. In effect, this rule functions similar in spirit to backprop in that it directly provides a (proxy) teaching signal that only affects the weights but not the activities of the neurons of a given layer. We found this to be a fast and effective rule for updating the parameters of blocks that contained stacks of operations of length greater than one, i.e., residual blocks, in order to achieve performance comparable to ResNet trained with backprop. (Note that one could, instead, utilize backprop to compute the local gradients/adjustments of the intermediate filters/activities of the block instead, offering an even more stable update scheme for within a block.)



\end{document}